\documentclass[10pt, conference, compsocconf]{IEEEtran}
\IEEEoverridecommandlockouts
\usepackage{cite}
\usepackage{caption}
\usepackage{color}
\usepackage{amsmath,amssymb,amsfonts}
\usepackage{algorithmic}
\usepackage{multirow}
\usepackage{booktabs}
\usepackage{subfigure}
\usepackage{graphicx}
\usepackage{textcomp}
\usepackage{xcolor}
\usepackage{xspace}
\newcommand{\tool}{{SCRIPT}\xspace}
\def\BibTeX{{\rm B\kern-.05em{\sc i\kern-.025em b}\kern-.08em
    T\kern-.1667em\lower.7ex\hbox{E}\kern-.125emX}}
\usepackage{hyperref}
\hypersetup{
    colorlinks=true,
    linkcolor=black,
    filecolor=black,      
    urlcolor=black,
    citecolor=black,
}
\begin{document}

\title{Source Code Summarization with Structural Relative Position Guided Transformer}

\author{\IEEEauthorblockN{Zi Gong\textsuperscript{1}, Cuiyun Gao\thanks{*Corresponding author}\textsuperscript{1}*, Yasheng Wang\textsuperscript{2}, Wenchao Gu\textsuperscript{3}, Yun Peng\textsuperscript{3}, Zenglin Xu\textsuperscript{1}*}
\IEEEauthorblockA{\textsuperscript{1}Department of Computer Science and Technology, Harbin Institute of Technology (Shenzhen), Shenzhen, China\\
\textsuperscript{2}Noah’s Ark Lab, Huawei Technologies, Shenzhen, China\\
\textsuperscript{3}Department of Computer Science and Engineering, The Chinese University of Hong Kong, Hong Kong, China\\
Email: gongzi@stu.hit.edu.cn, \{gaocuiyun, xuzenglin\}@hit.edu.cn,\\wangyasheng@huawei.com, \{wcgu, ypeng\}@cse.cuhk.edu.hk}
}

\maketitle

\captionsetup[figure]{name={Fig.},labelsep=period,singlelinecheck=off}
\captionsetup[table]{name={TABLE},labelsep=period,singlelinecheck=off}
\begin{abstract}
Source code summarization aims at generating concise and clear natural language descriptions for programming languages. Well-written code summaries are beneficial for programmers to participate in the software development and maintenance process. To learn the semantic representations of source code, recent efforts focus on incorporating the syntax structure of code into neural networks such as Transformer. Such Transformer-based approaches can better capture the long-range dependencies than other neural networks including Recurrent Neural Networks (RNNs), however, most of them do not consider the structural relative correlations between tokens, e.g., relative positions in Abstract Syntax Trees (ASTs), which is beneficial for code semantics learning.

To model the structural dependency, we propose a {\large S}tru{\large C}tural {\large R}elat{\large I}ve {\large P}osition guided {\large T}ransformer, named \tool. \tool first obtains the structural relative positions between tokens via parsing the ASTs of source code, and then passes them into two types of Transformer encoders. One Transformer directly adjusts the input according to the structural relative distance; and the other Transformer encodes the structural relative positions during computing the self-attention scores. Finally, we stack these two types of Transformer encoders to learn representations of source code. Experimental results show that the proposed \tool outperforms the state-of-the-art methods by at least 1.6\%, 1.4\% and 2.8\% with respect to BLEU, ROUGE-L and METEOR on benchmark datasets, respectively. We further show that how the proposed \tool captures the structural relative dependencies.
\end{abstract}

\section{Introduction}
Recently, utilizing deep learning techniques to tackle problems in software development and maintenance process becomes an active area of research. In particular, methods based on deep learning have achieved impressive performance in various tasks associated with source code, such as code search \cite{gu2018deep,wan2019multi,husain2019codesearchnet}, code clone detection \cite{tufano2018deep,zhao2018deepsim,ain2019systematic}, code summarization \cite{alon2018code2seq,mahmud2021code,wu2020code}, and code generation \cite{ahmad2021unified,xu2020incorporating,alon2020structural}. These studies have greatly facilitated programmers to work efficiently. In this work, we focus on the task of source code summarization. The goal of this task is to generate short natural language descriptions of source code, and provide a convenient way for programmers to quickly understand the semantics of source code.

A recent study showes that developers spend an average of 59$\%$ of their time on program understanding \cite{xia2017measuring}. It is no doubt that high-quality code summaries are key to improve the efficiency of program understanding \cite{he2019understanding}. However, high-quality or up-to-date summaries of source code are often absent from various software projects for a variety of reasons. Thus, many researchers have devoted to the works of automatic code summarization for decades. 
% \py{decades or a decade. make sure that this research has lasted for a decade.}

The key point of the task is to learn the semantic representations of source code by modeling the relation between source code and natural language. Early studies consider the task of code summarization as an information retrieval (IR) problem. For example, \cite{haiduc2010use,eddy2013evaluating,wong2015clocom} use heuristic rules or vector space model to identify critical words from source code and then synthesize the words into natural language descriptions. The IR-based methods focusing on the keyword extraction from source code while ignoring the natural semantic relations between source code and natural language. Thus, the IR-based methods would generate summaries with poor readability.

With the rapidly growing of deep learning and the availability of large-scale data, researchers discover some Seq2Seq models such as Recurrent Neural Networks (RNNs) \cite{cho2014learning,bahdanau2014neural} are capable of modeling the semantic relations between source code and summary. Iyer \textit{et al.} \cite{iyer2016summarizing} first propose to use Long Short-Term Memory (LSTM) networks with an attention mechanism to generate summaries of C\# and SQL code by learning representations from noisy datasets that extract from online programming websites. Subsequently, increasing works \cite{jiang2017automatically,loyola2017neural,liang2018automatic} generate source code summaries based on Seq2Seq models without considering the structural properties of code. To better capture the structural information, Hu \textit{et al.} \cite{hu2018deep} propose Structure-based Traversal (SBT) to attach structural information of code to input, and Shido \textit{et al.} \cite{shido2019automatic} propose Tree-LSTM to directly model structural representations of code. However, the RNN-based approaches may encounter bottleneck when modeling long sequences due to its poor long-term dependency \cite{wu2020code}. To capture the long-range dependencies, Ahmad \textit{et al.} \cite{ahmad2020transformer} propose a Transformer-based \cite{vaswani2017attention} method which combines copy mechanism \cite{see2017get} and relative position encoding (RPE) \cite{shaw2018self}. Most recently, SiT \cite{wu2020code} is proposed to incorporate multi-view graph matrix into Transformer's self-attention mechanism, and achieves the state-of-the-art performance.

\begin{figure}[htb]
    \centering
    \quad
    \subfigure[Example of source code and corresponding summaries.]{
    \includegraphics[width=8.5cm]{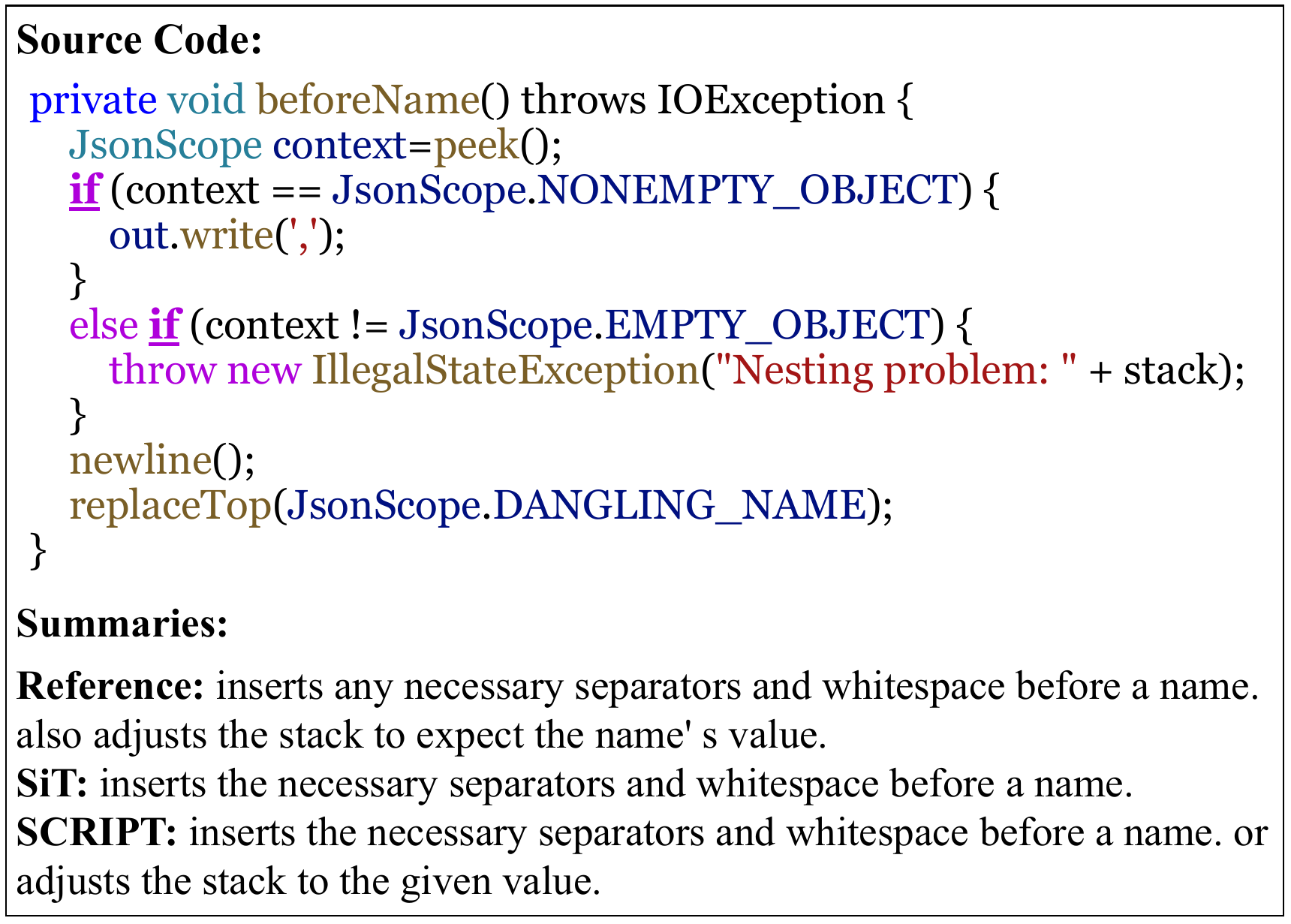}
    }
    \quad
    \subfigure[The AST of source code in (a), white boxes refer to type of the node, and green boxes refer to value of the node.]{
    \includegraphics[width=8.5cm]{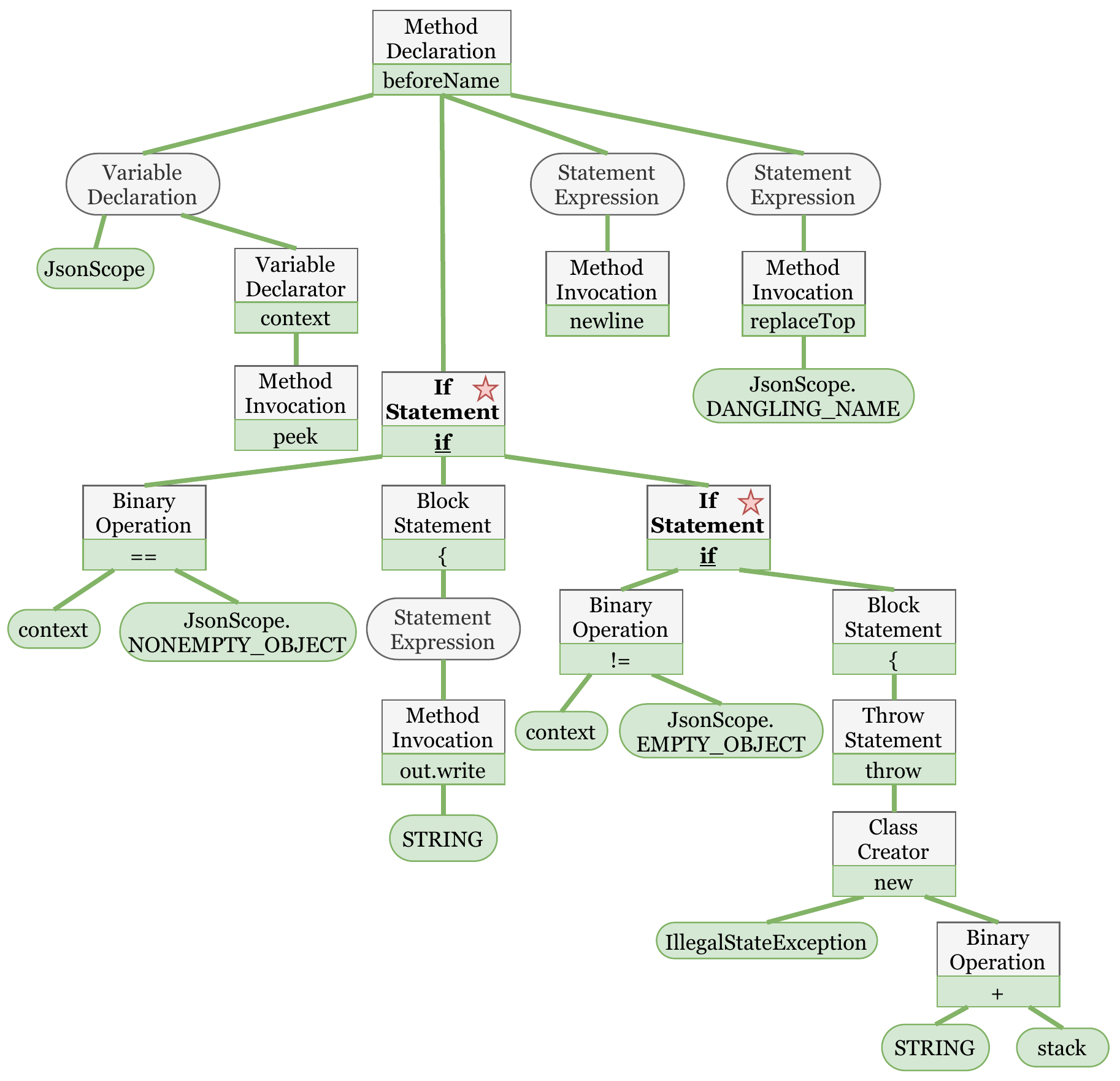}
    }
    \caption{An example of the source code with corresponding summaries and the AST.}
    \label{fig1}
\end{figure}

Although the Transformer-based methods show superior performance than other DL-based methods for the task, they only exploit the sequential relative positions between code tokens. In this work, we suppose that the structural relative positions between tokens (i.e., relative positions of nodes in ASTs) are also essential for learning code semantics. That is, tokens that are far from each other in the sequential input are likely to be very close in the corresponding AST. For example, as shown in Figure~\ref{fig1}(a), the distance between the both underlined tokens ``if" is far away in the sequential input, but they, i.e., the red-star marked nodes, are close in corresponding AST (only 1 hop), as shown in Figure~\ref{fig1}(b). By comparing the summaries generated by SiT \cite{wu2020code} and \tool in Figure~\ref{fig1}(a), we can observe that SiT does not well learn the relations between the two ``if" tokens, and produces relatively incomplete summary. With the information of structural relative positions involved, the proposed \tool can capture the semantics of the two ``if'' statements, and thus accurately predict the relatively complete code summary. More detailed analysis can be found in Section~\ref{subsec:vis}.

Specifically, \tool first extracts the structural relative positions between tokens from the ASTs of source code. Two types of Transformer encoders are then designed to generate structural representations of source code for the task. One Transformer directly adjusts the input to enhance structural dependencies according to the structural relative positions.
% \py{this phase is vague, what do you mean by saying directly adjust? maybe you want to say transform the token sequence orders into AST orders?}
For the example in Figure~\ref{fig1}, the dependency between the two underlined ``if"  tokens is weighted due to their close distance in the AST. The other Transformer built upon SiT \cite{wu2020code} innovatively incorporates the structural relative positions into the computation of the self-attention scores. In this way, the dependencies between the tokens that are close in AST can be further enhanced. The difference between the two types of Transformer is the way of learning structural information of code. We finally stack the two types of Transformer encoders as \tool encoder module.

We conduct experiments on two benchmark datasets in Java and Python, respectively. Experimental results demonstrate \tool outperforms the state-of-the-art solutions by at least 1.6\%, 1.4\% and 2.8\% with respect to BLEU, ROUGE-L and METEOR on benchmark datasets, respectively. Extensive human evaluations further endorse the validity of the summaries generated by our approach.

In summary, the main contributions of this paper are outlined as follows:
\begin{itemize}
    \item We propose two novel types of Transformer encoders to capture the structural relative positions between tokens for better learning code semantics.
    \item We propose \tool which stacks the two types of Transformer encoders to learn representation of source code for code summarization.
    \item Extensive experiments, including qualitative experiments and human evaluations, verify the effectiveness of the proposed \tool. Our source code is publicly released at \url{https://github.com/GoneZ5/SCRIPT}.
\end{itemize}

The rest of the paper is organized as follows. Section presents an overview of our approach and describes the details of each component. Section III presents the experimental settings. Section IV shows the experimental results. Section V presents discussion about some results and the threats to validity. Section VI presents related works of source code summarization. Finally, we conclude our work and present future work in Section VII.
\section{Proposed Approach}
\begin{figure}[htb]
    \centering
    \includegraphics[width=8.5cm]{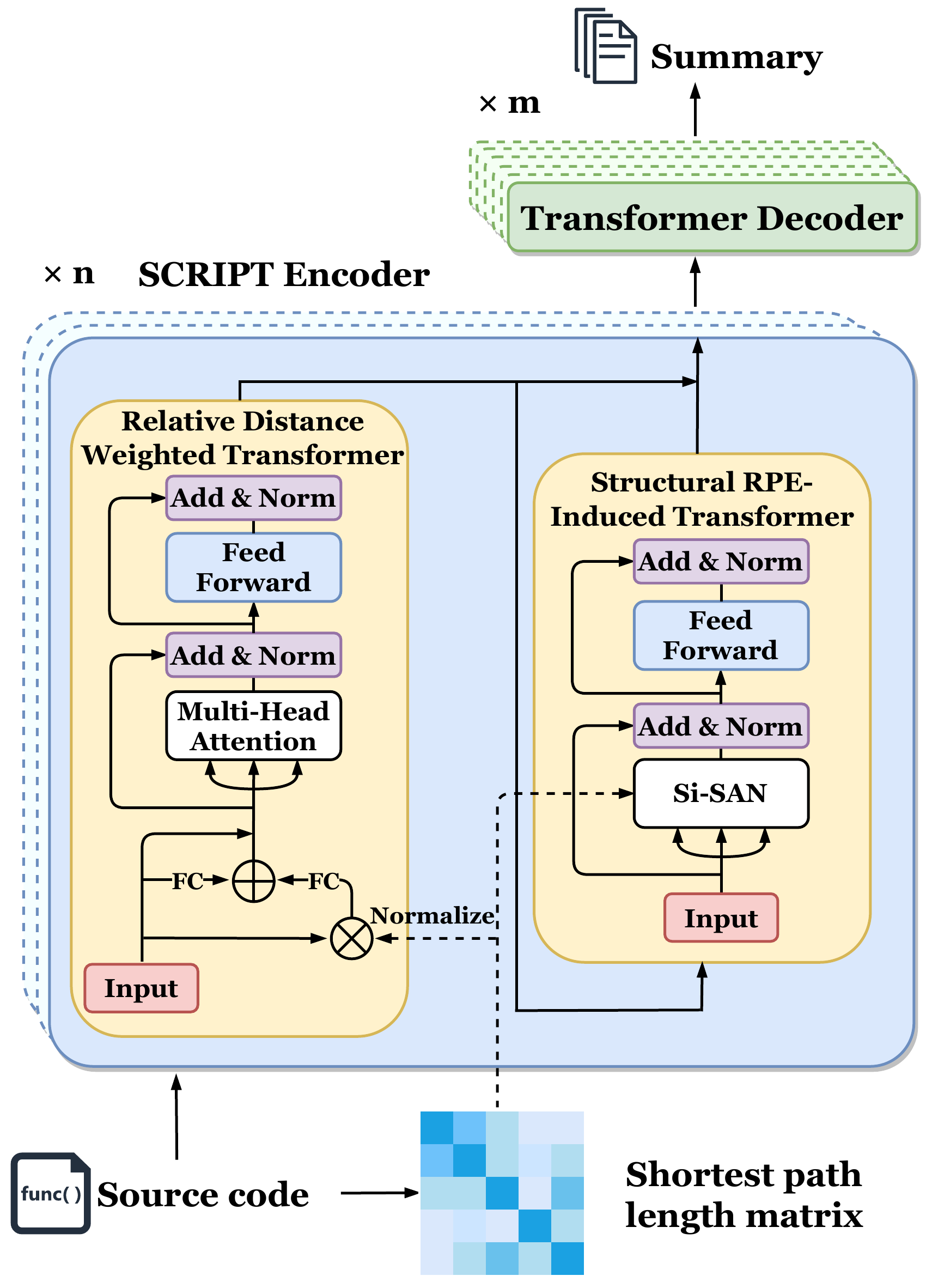}
    \caption{Overall architecture of the proposed \tool. The ``Si-SAN'' in the right Structural RPE-Induced Transformer refers to the Structure-induced Self-Attention Networks which is proposed by Wu \textit{et al.} \cite{wu2020code}.}
    \label{fig2}
\end{figure}

In this section, we first present an overview of our model. Then we describe the detailed components of \tool in the following sub-sections. Figure~\ref{fig2} shows the overall architecture of \tool. It takes token sequences of source code and extracted structural information as inputs. Two types of Transformer encoders compose the encoder part of \tool, including Relative Distance Weighted Transformer and Structural RPE-induced Transformer. They help model to well learn structural relative position knowledge. We pass the inputs into \tool encoder. For decoder, we use the original Transformer decoder with mask multi-head attention and cross attention. Finally, the corresponding summary of input code is generated by autoregressive decoding. 

\subsection{Relative Distance Weighted Transformer}
The vanilla Transformer encoder~\cite{vaswani2017attention} initially treats the relations between any tokens equally and learns the ``real'' relationships implicitly through training. If we leverage the relative distance between nodes in the AST to explicitly weight the structural dependencies, the model would learn the program semantics correlation between the two tokens. Intuitively, the distance between two tokens in the AST of source code may reflect the relevance in program semantics correlation, i.e., distance is inversely proportional to semantics correlation. Therefore, we introduce the shortest path length of nodes in the AST, which represents the structural relative positions between tokens. Then we construct the Relative Distance Weighted Transformer (RDW-Transformer) as shown in the left part of Transformer Encoder in Figure~\ref{fig2}.

\begin{figure}[htb]
    \centering
    \subfigure[AST.]{
    \includegraphics[width=3cm]{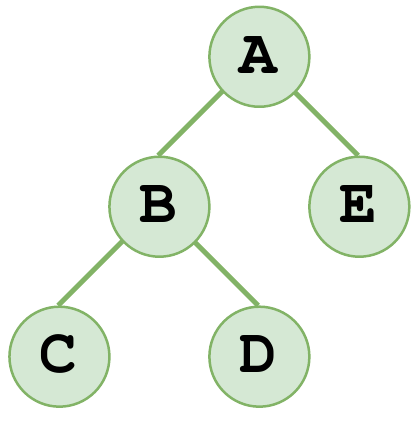}
    }
    \quad
    \hspace{0.8cm}
    \subfigure[Shortest path length matrix.]{
    \includegraphics[width=3cm]{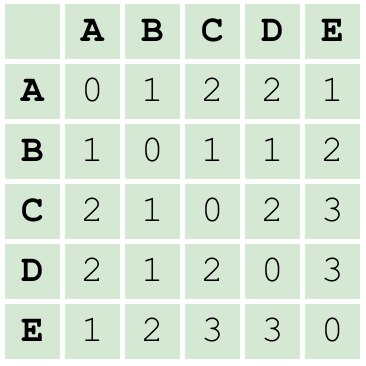}
    }
    \caption{Case of AST and corresponding shortest path length matrix.}
    \label{fig3}
\end{figure}

First, we treat the AST of source code as an undirected graph, since it is indispensable to ensure accessibility at a finite distance of each two nodes. Then we consider the shortest path lengths between nodes in the AST as structural relative positions. We represent the shortest path length of the AST as a structural relative position matrix $M$ as shown in Figure~\ref{fig3}(b). Such a representation has a benefit that the position matrix $M$ can be easily integrated into Transformer without changing the architecture of Transformer. Specifically, we define the shortest path length in the AST between the token $i$ and token $j$ as $\operatorname{spl}(i, j)$. Given an AST of source code snippet, we define the element ${M}_{i j}$ of position matrix $M$ in $i$-th row and $j$-th column as:

\begin{equation}
    \boldsymbol{M}_{i, j}=\left\{
    \begin{array}{cl}
        \operatorname{spl}(i, j), & i \neq j,\\
        0, & i = j.
    \end{array}\right.
    \label{eqa14}
\end{equation}

We follow the view that the closer two tokens are in the AST of source code, the stronger their semantics correlation is. Then we normalize the matrix $M$ after taking the reciprocal for all non-zero elements. The normalized position matrix $\bar{M}$ is formulated as:

\begin{equation}
    \boldsymbol{\bar{M}}_{i, j}=\left\{
    \begin{array}{cl}
        \frac{1 / M_{i, j}}{\sum_{z \in\left\{y \mid M_{i, y} \neq 0\right\}}\left(1 / M_{i, z}\right)}, & \text { if } \boldsymbol{M}_{i, j} \neq 0, \\
        0, & \text { otherwise. }
    \end{array}\right.
    \label{eqa15}
\end{equation}

Given the original input of \textit{n}-th layer of Transformer encoder $H^{n-1}$, i.e., output of the previous layer, we fuse the structural position matrix $\bar{M}$ into the original input and get the structural distance weighted representation as:

\begin{equation}
    \widehat{H}^{n}=\sigma\left(\operatorname{FC}^{1}_{n}\left(H^{n-1}\right)+\operatorname{FC}^{2}_{n}\left(\bar{M} H^{n-1}\right)\right),
    \label{eqa16}
\end{equation}
where $\sigma$ refers to the sigmoid activation function, $\operatorname{FC}^{1}_{n}$ and $\operatorname{FC}^{2}_{n}$ refer to fully-connected layers in the $n$-th layer of the Transformer encoder. The expression $\bar{M} H^{n-1}$ contributes in weighting structural dependencies between tokens according to structural relative position information provided by the normalized matrix $\bar{M}$. For the closer tokens in the AST, the dependencies between them are weighted more, and vice verse. By this way, we can directly provide the program semantics correlations between tokens to RDW-Transformer. 

Finally, we incorporate the structural distance weighted representation $\widehat{H}^{n}$ into the original input $H^{n-1}$, which is served as the final input:

\begin{equation}
    H^{n}=\operatorname{Transformer}_{n}\left(\operatorname{Aggr}\left(H^{n-1}, \widehat{H}^{n}\right)\right),
    \label{eqa17}
\end{equation}
where $H^{n}$ is the output of the current Transformer encoder layer. Each Transformer layer $\operatorname{Transformer}_{n}$ contains an architecturally identical transformer block, which consists of a multi-head self-attention $\operatorname{MultiAttn}$ and a feed forward network $\operatorname{FFN}$.

\subsection{Structural RPE-Induced Transformer}
\textbf{Self-Attention} Vaswani \textit{et al.} \cite{vaswani2017attention} propose self-attention which can be generally formulated as:

\begin{equation}
    \operatorname{Attention}(Q, K, V)=\operatorname{softmax}\left(\frac{Q K^{T}}{\sqrt{d_{k}}}\right) V,
    \label{eq1}
\end{equation}
where queries, keys and values are packed together into matrices $Q$, $K$ and $V$ respectively, and $d_{k}$ is the dimension of $K$.

\textbf{Sequential Relative Positional Encoding} Shaw \textit{et al.} \cite{shaw2018self} propose an extended transformer model that considers the pairwise relationships between input elements. In their approach, they add the sequential relative position representations to the projection of $K$ and $V$. They not only consider relative position relationships between tokens as calculating self-attention scores, but also propagate the relationships in computing context vectors process. Furthermore, Shaw \textit{et al.} \cite{shaw2018self} hypothesized that relative position information is not useful beyond a certain distance. They define the maximum relative position as a constant $k$ and clip the maximum distance to generalize the model.

\textbf{AST Relative Positional Encoding} Although sequential relative positional encoding can capture the relative positional dependencies between serialized tokens, it neglects to model the structural relative positional correlations. To this end, we utilize the shortest path length of nodes in AST again. Then we explain how to encode AST relative positions into Transformer.

We follow the idea of Shaw \textit{et al.} \cite{shaw2018self} to encode structural relative positions. We define the sequential relative position representations between $\boldsymbol{x}_{i}$ and $\boldsymbol{x}_{j}$ (i.e. the $i$-th and $j$-th words) in an input sequence as $\boldsymbol{a}_{i j}^{V}, \boldsymbol{a}_{i j}^{K} \in \mathbb{R}^{d_{k}}$, and the structural relative position embedding between node $n_{i}$ and node $n_{j}$ in AST as $\boldsymbol{b}_{i j}^{V}$, $\boldsymbol{b}_{i j}^{K} \in \mathbb{R}^{d_{k}}$. The sequential and structural relative position representations are added to the projection of $V$ to propagate both types of relative positional information as:

\begin{equation}
    \boldsymbol{z}_{i}=\sum_{j=1}^{n} \alpha_{i j}\left(\boldsymbol{x}_{j} W^{V}+\boldsymbol{a}_{i j}^{V}+\boldsymbol{b}_{i j}^{V}\right).
    \label{eqa7}
\end{equation}

We also add the two types of relative position representations to the projection of $K$ in order to consider relative position relationships between tokens in calculating $e_{i j}$ as:

\begin{equation}
    e_{i j}=\frac{\boldsymbol{x}_{i} W^{Q}\left(\boldsymbol{x}_{j} W^{K}+\boldsymbol{a}_{i j}^{K}+\boldsymbol{b}_{i j}^{K}\right)^{T}}{\sqrt{d_{z}}}.
    \label{eqa8}
\end{equation}

Similar to settings of Shaw \textit{et al.} \cite{shaw2018self}, we also assume that when the shortest path length beyond a certain threshold, the structural relative position information is not useful. Thus, we define the maximum distance to a constant ${l}$ and introduce the operation of clipping to limit the maximum distance. We consider $l + 1$ unique structural relative position labels as:

\begin{equation}
    \boldsymbol{b}_{i j}^{V}=\boldsymbol{w}_{\operatorname{clip}({d}_{i j}, l)}^{V},
    \label{eqa9}
\end{equation}
\begin{equation}
    \boldsymbol{b}_{i j}^{K}=\boldsymbol{w}_{\operatorname{clip}({d}_{i j}, l)}^{K},
    \label{eqa10}
\end{equation}
\begin{equation}
    \operatorname{clip}(x, l)=\min (x, l),
    \label{eq6}
\end{equation}
where ${d}_{i j}$ is the shortest path length between node ${n}_{i}$ and ${n}_{j}$ in AST.

\textbf{Structural RPE-induced Transformer} To better encode the structural relative position in AST, we propose Structural RPE-induced Transformer (SRPEi-Transformer) as shown in the right part of Transformer Encoder of Figure~\ref{fig2}.

Specifically, we follow the method of Wu \textit{et al.} \cite{wu2020code} to extract multi-view graph matrix based on different adjacency matrices of code semantics, which are abstract syntax matrix ($A_{ast}$), control flow matrix ($A_{fl}$) and data dependency matrix ($A_{dp}$). We formulate it as:

\begin{equation}
    A_{mv} = \alpha A_{ast} + \beta A_{fl} + \gamma A_{dp},
\end{equation}
where $\alpha$, $\beta$, $\gamma$ refer to the corresponding weight for each code semantics matrix. $A_{mv}$ is able to mask redundant attention and enhance model robustness. We formulate the structural RPE-induced self-attention network (SRPEi-SAN) as:

\begin{equation}
    \operatorname{SRPEi-SAN}(X)=\operatorname{Softmax}\left(\frac{A_{mv} \cdot Q K^{T}}{\sqrt{d_{k}}}\right) V,
    \label{eqa11}
\end{equation}
where $X=\left(x_{1}, \ldots, x_{s}\right)$ refers to the representation of input sequence, $s$ refers the length of sequence and ${d}_{k}$ refers the dimension per attention head. And the Q, K and V are the projections of queries, keys and values respectively.

Then we incorporate AST relative positional encoding into the calculation process of self-attention scores and context vectors. In detail, we add the AST relative positional representations ${b}^{V}$ and ${b}^{K}$ to $V$ and $K$ respectively as:

\begin{equation}
    V=\left(X W^{V}+a^{V}+b^{V}\right),
    \label{eq12}
\end{equation}
\begin{equation}
    K=\left(X W^{K}+a^{K}+b^{K}\right).
    \label{eq13}
\end{equation}

Then we connect a feed forward layer $\operatorname{FFN}$ to the back of SRPEi-SAN to build SRPEi-Transformer. In this way, SRPEi-Transformer can simultaneously capture both of textual and structural relative positional dependencies for learning more informative representations of source code. It is worth noting that we only equip SRPEi-Transformer encoder layers with AST relative positional encoding, rather than the other Transformer encoder layers. The details about this will be presented in Section~\ref{subsec:depl}.

\subsection{\tool Encoder Module}
In order to take advantage of both Transformer encoders, we introduce \tool encoder module, which is a stack of the two types of Transformers, i.e., RDW-Transformer and  SRPEi-Transformer. In the encoder module, RDW-Transformer layer is followed by SRPEi-Transformer layer, and the output of the module $\bar{H}$ is the combination of both layers' output:

\begin{equation}
    \bar{H} = \operatorname{Aggr}(H, H'),
    \label{eq14}
\end{equation}
where $H$ refers to the output of RDW-Transformer, and $H'$ refers to the output of SRPEi-Transformer, and we use a simple position-wise sum as the aggregation.

In our experimentation, we set the proposed model with 3 stacks of \tool encoder modules, i.e., 6 encoder layers, and 6 decoder layers.
\section{Experiments Setup}
\subsection{Baselines}
\begin{itemize}
    \item \textbf{CODE-NN} \cite{iyer2016summarizing} follows LSTM-based encoder-decoder architecture with attention mechanism, whose encoder processes source code snippets to context vectors and decoder decodes them to natural language descriptions.
    \item \textbf{Tree2Seq} \cite{eriguchi2016tree} is an end-to-end syntactic NMT model which directly uses a tree-based LSTM as an encoder. It extends a Seq2Seq model with the source code structure.
    \item \textbf{RL+Hybrid2Seq} \cite{wan2018improving} incorporates ASTs and sequential content of code snippets into a deep reinforcement learning framework. The basic architecture is a multi-encoder NMT, which jointly learn structural and sequential information by encoding ASTs and tokens of source code. It further uses reinforcement learning to solve the exposure bias problem during decoding and obtains better performance.
    \item \textbf{DeepCom} \cite{hu2018deep} flattens the AST into a sequence as input, which is obtained via traversing the AST with a structure-based traversal (SBT) method. This method introduces structural information of source code without changing the framework of model and the sequential form of inputs.
    \item \textbf{API+Code} \cite{hu2018summarizing} utilizes a multi-encoder neural architecture. It encodes API sequence along with code token sequence, then generates summary from source code with transferred API knowledge. Essentially, the model trains an API sequence encoder by using an external dataset so that it can learn more abundant representations of source code.
    \item \textbf{Dual Model} \cite{wei2019code} treats code summarization (CS) and code generation (CG) as a dual task. It trains the two tasks jointly by a dual training framework to simultaneously improve the performance of CS and CG tasks.
    \item \textbf{Transformer} \cite{ahmad2020transformer} is a Transformer-based model with relative positional encoding and copy mechanism, which is effective in capturing long-range dependencies of source code.
    \item \textbf{SiT} \cite{wu2020code} incorporates multi-view graph matrix into Transformer’s self-attention mechanism. Essentially, it improves the performance via masking redundant attention in the calculation process of self-attention scores.
\end{itemize}

\subsection{Evaluation Metrics}
We evaluate the source code summarization performance by using three metrics, BLEU-4 \cite{papineni2002bleu}, ROUGE-L \cite{lin2004rouge} and METEOR \cite{banerjee2005meteor}, which all measure the quality of text generation.

\textbf{BLEU} is a standard evaluation metric in the source code summarization works. BLEU score calculates the geometric mean of $n$-gram matching precision scores, which is multiplied by a brevity penalty to prevent very short generated sentence. In a nut shell, BLEU can be thought of as a precision score as: how much of the generated text appears in the reference text. In this paper, we choose the BLEU-4 as our metric.

\textbf{ROUGE-L} is used primarily in text summarization tasks in the NLP literature. The score allows multiple references since there may be multiple correct summaries of a text. The metric is defined as the length of longest common subsequence between generated sentence and reference, and based on recall scores. ROUGE-L can be thought of as a recall score as: how much of the reference appears in the generated text.

\textbf{METEOR} is a evaluation metric for machine translation task which is we treat source code summarization as. The metric is based on the harmonic mean of unigram precision and recall, with recall weighted higher than precision. METEOR can produce good correlation with human judgement at the sentence or segment level.

\subsection{Datasets and Pre-processing}
In this sub-section, we present the sources of our experiment datasets and the methods of data pre-processing.

\renewcommand\arraystretch{1.2}
\begin{table}[htb]\normalsize
    \centering
    \setlength{\tabcolsep}{4mm}{
    \begin{tabular}{l r r} \bottomrule
        Datasets & Java & Python \\
        \hline
        Train & 69,708 & 55,538 \\
        Validation & 8,714 & 18,505 \\
        Test & 8,714 & 18,502 \\
        \hline
        Avg. tokens in code & 73.76 & 49.42 \\
        Avg. tokens in summary & 17.73 & 9.48 \\
        \bottomrule
    \end{tabular}
    }
    \caption{Statistics of the experimental datasets. We follow the method of Wei \textit{et al.} \cite{wei2019code} to split the Python dataset, and the Java dataset splits are publicly available.}
    \label{tab:2}
\end{table}

\textbf{Datasets.} We conduct the source code summarization experiments on two public parallel datasets, including Java \cite{hu2018summarizing} and Python \cite{wan2018improving}. The statistics of the two datasets are shown in Table~\ref{tab:2}. The Java dataset is collected from GitHub\footnote{\url{https://github.com/}}, which contains Java methods and summaries extracted from Java projects from 2015 to 2016. We pre-process the Java dataset following Wu \textit{et al.} \cite{wu2020code} and each data sample is organized as a pair of $<$code, summary$>$. The original Python dataset is collected by Barone\textit{et al.} \cite{barone2017parallel}, consisting of about 110K parallel data samples. Note that we do not follow the method of Wu \textit{et al.} \cite{wu2020code} for cleaning and splitting the Python dataset, but follow the way of Wei \textit{et al.} \cite{wei2019code}. The main reason is that we cannot directly parse the ASTs from the tokenized Python dataset of Wu \textit{et al.} \cite{wu2020code}.

\textbf{Shortest path length of AST.} We first use \textit{javalang}\footnote{\url{https://github.com/c2nes/javalang}} module of Python to parse Java code and then fetch ASTs in a dictionary form, while we use \textit{asttokens} module provided by Wu \textit{et al.} \cite{wu2020code} to extract ASTs for Python dataset. Then we traverse the ASTs according to SBT \cite{hu2018deep} to get adjacency matrix of ASTs. Finally, we utilize the Floyd algorithm \cite{floyd1962algorithm} to compute the shortest path length of ASTs.

\textbf{Out-Of-Vocabulary.} The vast operators and identifiers in program language may produce much larger vocabulary than natural language, which can cause Out-of-Vocabulary problem. To avoid this problem, we apply $CamelCase$ and $snake\_case$ tokenizers that are consistent with Ahmad \textit{et al.} \cite{ahmad2020transformer} to reduce the vocabulary size of source code. In addition, we replace all string and number identifiers with STR and NUM respectively on both benchmarks, which can further decrease the source-side vocabulary size.

\subsection{Training details}
The implementation of \tool is based on the PyTorch\footnote{\url{https://pytorch.org//}} implementation of open-NMT\footnote{\url{https://github.com/OpenNMT/OpenNMT-py}}. The word embedding size of source code and summary are set to 512. We set the batch size to 32 and use the Adam \cite{kingma2014adam} optimizer with an initial learning rate in \{5e-5, 1e-4\}, and set the warm-up rate of 0.06 and L2 weight decay of 0.01. To avoid over-fitting, we adopt dropout \cite{srivastava2014dropout} with a drop probability in \{0.2, 0.4\}. We train \tool and SiT \cite{wu2020code} for a maximum of 200 epochs, and if validation performance does not improve over 20 consecutive iterations, an early stop is performed. We use a beam search during inference and set the beam size to 5 for choosing the best result.
\section{Experimental Results}
We present the experimental results and analysis through the following research questions.

\subsection{${RQ}_{1}:$ How does \tool perform compared to the baselines?}

\renewcommand\arraystretch{1.2}
\begin{table*}[htb]\normalsize
    \centering
    \setlength{\tabcolsep}{2.8mm}{
    \begin{tabular}{l ccc ccc} \bottomrule
        \multirow{2}{*} {\textbf{Model}} & \multicolumn{3}{c} {\textbf{Java}} & \multicolumn{3}{c} {\textbf{Python}} \\
        \cline {2-7} & \textbf{BLEU} & \textbf{ROUGE-L} & \textbf{METEOR} & \textbf{BLEU} & \textbf{ROUGE-L} & \textbf{METEOR} \\
        \hline 
        CODE-NN (Iyer et al., 2016) & 27.60 & 41.10 & 12.61 & 17.36 & 37.81 & 09.29 \\
        Tree2Seq (Eriguchi et al., 2016) & 37.88 & 51.50 & 22.55 & 20.07 & 35.64 & 08.96 \\
        RL+Hybrid2Seq (Wan et al., 2018) & 38.22 & 51.91 & 22.75 & 19.28 & 39.34 & 09.75 \\
        DeepCom (Hu et al., 2018a) & 39.75 & 52.67 & 23.06 & 20.78 & 37.35 & 09.98 \\
        API + Code (Hu et al., 2018b) & 41.31 & 52.25 & 23.73 & 15.36 & 33.65 & 08.57 \\
        Dual Model (Wei et al., 2019) & 42.39 & 53.61 & 25.77 & 21.80 & 39.45 & 11.14 \\
        Transformer (Ahmad et al., 2020) & 44.58 & 54.76 & 26.43 & 32.52 & 46.73 & 19.77 \\
        \bottomrule
        SiT* (Wu et al., 2021) & 45.70 & 55.54 & 27.55 & 33.46 & 47.50 & 20.28 \\
        \tool & \textbf{46.89} & \textbf{56.69} & \textbf{28.48} & \textbf{34.00} & \textbf{48.15} & \textbf{20.84} \\
        \tool w/o RDW-Transformer & 46.72 & 56.50 & 28.32 & 33.75 & 47.66 & 20.65 \\
        \tool w/o SRPEi-Transformer & 46.42 & 56.29 & 28.15 & 33.60 & 47.61 & 20.43 \\
        \bottomrule
    \end{tabular}
    }
    \caption{Comparison of the performance of our method with other baseline methods on Java and Python benchmarks in terms of BLEU, ROUGE-L and METEOR. * refers to models we rerun. We obtain the results of upper part from Wu \textit{et al.} \cite{wu2020code} reporting. Note that we only rerun SiT, since the Python dataset it uses is different from ours and the upper part approaches. The last two lines is the ablation studies about the two components mentioned ahead. We show that our approach are even stronger than all baselines.}
    \label{tab:3}
\end{table*}

The rationale for this research question is to compare \tool with other baselines for the task of source code summarization.

Table~\ref{tab:3} shows the overall results of different methods for generating source code summaries on Java and Python benchmarks in terms of the three automated evaluation metrics. We calculate the values of the metrics following the same scripts used by SiT \cite{wu2020code}. For these evaluation metrics, the larger value indicates the better performance, and we mark the best one of each metric in bold.

From the table, SiT is strong enough as it outperforms all the previous works by a significant margin. However, \tool is more powerful than SiT and achieves more impressive performance. On Java dataset, \tool improves the performance of BLEU, ROUGE-L and METEOR by 2.6\%, 2.1\% and 3.4\% respectively compared with SiT. \tool also boosts SiT by 1.6\%, 1.4\% and 2.8\% on BLEU, ROUGE-L and METEOR respectively on the Python dataset. The results demonstrate that the structural relative position information is significantly important for model to well learn code semantics for the task of source code summarization.

Specifically, \tool achieves more noticeable improvements on Java than Python, increasing by 1.19, 1.15 and 0.93 points on BLEU, ROUGE-L and METEOR respectively compared with SiT. A possible reason may be the size of source-side vocabulary. After tokenizing and replacing all string and number identifiers with STR and NUM respectively on two datasets, Java's vocabulary size (102726) is larger than Python's (46193). This difference makes it much more challenging on Java. Thus, the superiority of \tool on Java tends to be notable.

\subsection{${RQ}_{2}:$ How effective are the main components of \tool?}

In this paper, we design two design two types of Transformer encoders that are able to explicitly learn structural relative position information. The rationale for this research question is to know whether they are effective. The ablation experiment results as shown in the last two lines of Table~\ref{tab:3}.

In this experiment, we first train \tool without RDW-Transformer, i.e., replacing all RDW-Transformer encoders with normal Transformer encoders; we also train \tool without SRPEi-Transformer, i.e., replacing all SRPEi-Transformer encoders with normal Transformer encoders. On the one hand, comparing with our full model, the dropping performance demonstrates that: (1) RDW-Transformer is indispensable for \tool to attach structural relative positional information to input of Transformer encoder, and (2) SRPEi-Transformer is capable of helping model to capture structural relative positional dependencies through encoding AST relative positions. On the other hand, comparing with SiT, the significantly boosting of performance indicates that both of two types of Transformer encoder can help model to learn more structural representations of source code, and assist decoder in generating higher-quality summaries.

For our full model, the removal of any type of Transformer results in performance dropping. The last two lines of Table~\ref{tab:3} shows that the removal of SPREi-Transformer drops more performance than the removal of RDW-Transformer. This discovery indicates the SPREi-Transformer play a more important role than RDW-Transformer in encouraging model to learn structural relative positional information.

\subsection{${RQ}_{3}:$ How the deployed location of AST relative positional encoding make effects on performance?}\label{subsec:depl}

\renewcommand\arraystretch{1.2}
\begin{table*}[htb]\normalsize
    \centering
    \setlength{\tabcolsep}{2.5mm}{
    \begin{tabular}{l ccc ccc} \bottomrule
        \multirow{2}{*} {\textbf{Model}} & \multicolumn{3}{c} {\textbf{Java}} & \multicolumn{3}{c} {\textbf{Python}} \\
        \cline {2-7} & \textbf{BLEU} & \textbf{ROUGE-L} & \textbf{METEOR} & \textbf{BLEU} & \textbf{ROUGE-L} & \textbf{METEOR} \\
        \hline 
        Equip in RDW-Transformer & 46.53 & 56.36 & 28.25 & 33.68 & 47.70 & 20.52 \\
        Equip in SiT & \textbf{46.89} & \textbf{56.69} & \textbf{28.48} & \textbf{34.00} & \textbf{48.15} & \textbf{20.84} \\
        Equip in all encoder layers & 46.66 & 56.45 & 28.33 & 33.70 & 47.81 & 20.64 \\
        \bottomrule
    \end{tabular}
    }
    \caption{Results of contrastive experiments about encoding AST relative positions in different types of Transformer encoder. Note that \tool is the model corresponding to the second line in this table.}
    \label{tab:4}
\end{table*}

The rationale for this research question is to explore the applicability of AST relative positional encoding, and to figure out the intrinsic relevance among AST relative positional encoding, SRPEi-Transformer and RDW-Transformer. Therefore, we conduct a set of contrastive experiments about encoding AST relative positions in different types of Transformer encoder. We use the variant of SiT \cite{wu2020code}, which replacing all normal Transformer encoder with RDW-Transformer encoder as the backbone. And we can deploy each type of Transformer encoder with AST relative positional encoding. Specifically, we add the relative position representations to the projections of keys and values in the calculation process of self-attention scores. We list three settings of contrastive experiments as follows:
\begin{itemize}
    \item[1.] Equip RDW-Transformer encoders with AST relative positional encoding.
    \item[2.] Equip SRPEi-Transformer encoders with AST relative positional encoding.
    \item[3.] Equip all Transformer encoders with AST relative positional encoding.
\end{itemize}

Table~\ref{tab:4} shows the result of contrastive experiment. The comparison of line 1 and line 2 demonstrates that SRPEi-Transformer is more suitable for AST relative positional encoding than RDW-Transformer. And when we just encode AST relative positions in SRPEi-Transformer encoders, the model achieves the best performance. By comparing line 1 and line 3, we observe that when we remove AST relative positional encoding from SRPEi-Transformer, the performance drops slightly. The two observations validate the AST relative positional encoding and the SRPEi-Transformer can be effectively combined to learn better code semantics. However, comparing line 2 and line 3, when we add the AST relative position representations to RDW-Transformer, the performance drops significantly. This observation indicates that the combination of the AST relative positional encoding and the RDW-Transformer fails to help model to learn better representations of code, even hurts the performance.

\subsection{Human Evaluation}
\renewcommand\arraystretch{1.2}
\begin{table}[htb]\normalsize
    \centering
    \setlength{\tabcolsep}{4mm}{
    \begin{tabular}{l|ccc}
        \hline Dataset & Metrics & SiT & Ours \\
        \hline \multirow{3}{*}{\textbf{Java}} 
        & Relevance & 3.17 & \textbf{3.61} \\
        & Similarity & 3.07 & \textbf{3.44} \\
        & Naturalness & 3.90 & \textbf{4.16} \\
        \hline \multirow{3}{*}{\textbf{Python}} 
        & Relevance & 3.02 & \textbf{3.45} \\
        & Similarity & 2.81 & \textbf{3.35} \\
        & Naturalness & 3.85 & \textbf{4.14} \\
        \hline
    \end{tabular}
    }
    \caption{Human evaluation results on Java and Python benchmark datasets.}
    \label{tab:5}
\end{table}
Although the above auto evaluation metrics can calculate the textual difference between generated summaries and references, they fail to truly reflect the semantic similarity. In addition, we perform human evaluations on the two benchmark datasets in terms of three metrics, including \textit{relevance} (degree of how generated summaries are relevant with source code snippets), \textit{similarity} of generated summaries and references, and \textit{naturalness} (grammaticality and fluency of the generated summaries). We only compare with SiT as it is the strongest baseline as depicted in Table~\ref{tab:3}. Specifically, we recruit 10 volunteers to evaluate the generated summaries, 3 PhDs and 7 Masters, all of them have at least 3-5 years of programming experience. We randomly select 100 source code samples and corresponding set of summaries (including generated by \tool, generated by SiT and ground-truth), where 50 Java samples and 50 Python samples. The 100 samples are then evenly divided into two groups, each group contains 25 Java samples and 25 Python samples. The volunteers are given the reference and two summaries that generated by our \tool and SiT in a random order. The results indicate that the summaries generated by \tool are better, which manifest in more relevant with the source code, more similar with the ground-truth and more grammatically fluent.
\section{Discussion}
In this section, we further compare our proposed approach and the best baseline SiT. Then we discuss situations where our method performs well. We finally present threats to validity in this work.

\subsection{Performance for Source Code and Summary of Different Lengths}

\begin{figure}[htb]
    \centering
    \subfigure[BLEU]{
    \includegraphics[width=4.2cm]{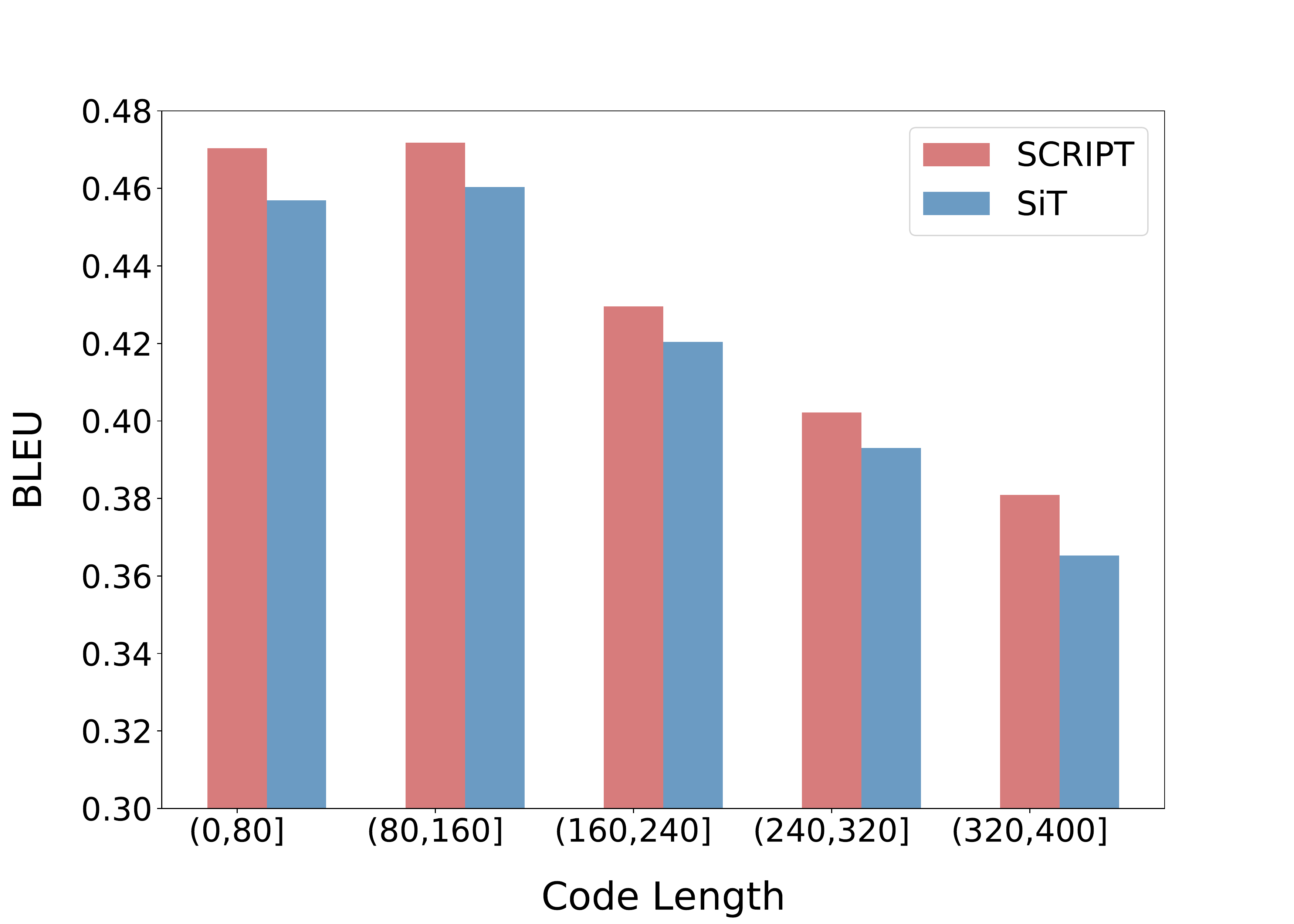}
    }
    \quad
    \hspace{-0.8cm}
    \subfigure[ROUGE-L]{
    \includegraphics[width=4.2cm]{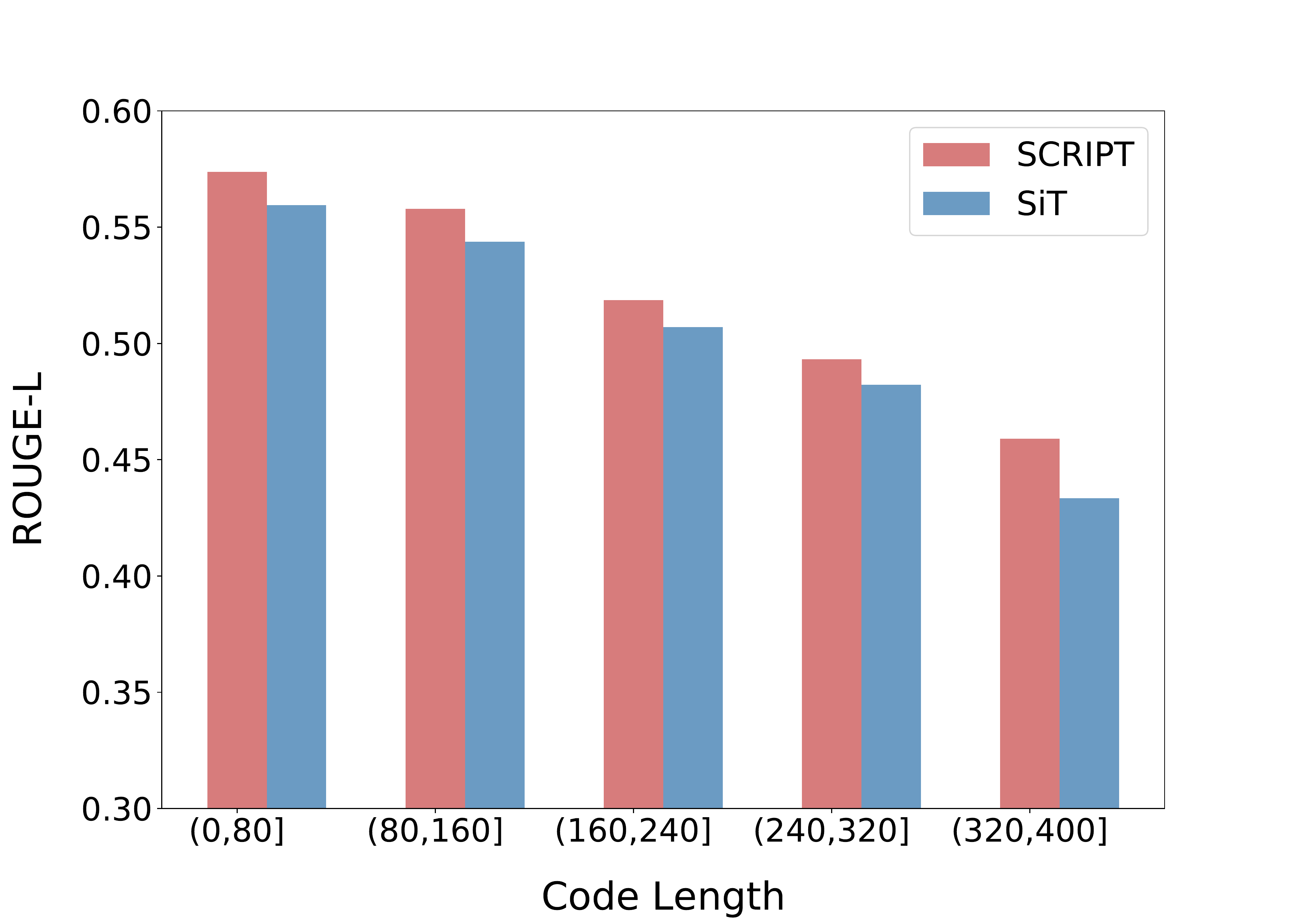}
    }
    \caption{Experimental results of our proposed method and SiT on Java dataset of different code lengths.}
    \label{fig4}
\end{figure}

\begin{figure}[htb]
    \centering
    \subfigure[BLEU]{
    \includegraphics[width=4.2cm]{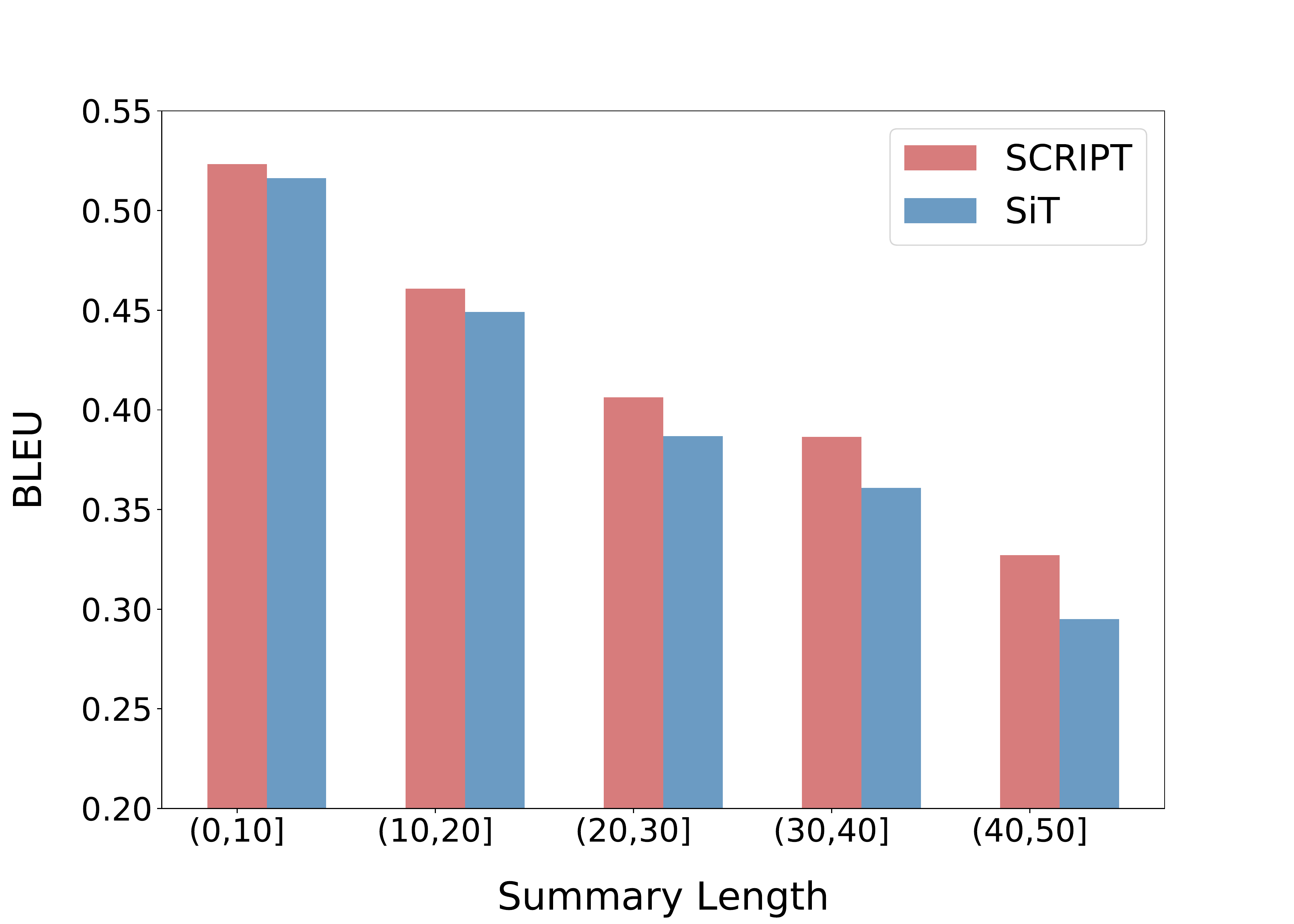}
    }
    \quad
    \hspace{-0.8cm}
    \subfigure[ROUGE-L]{
    \includegraphics[width=4.2cm]{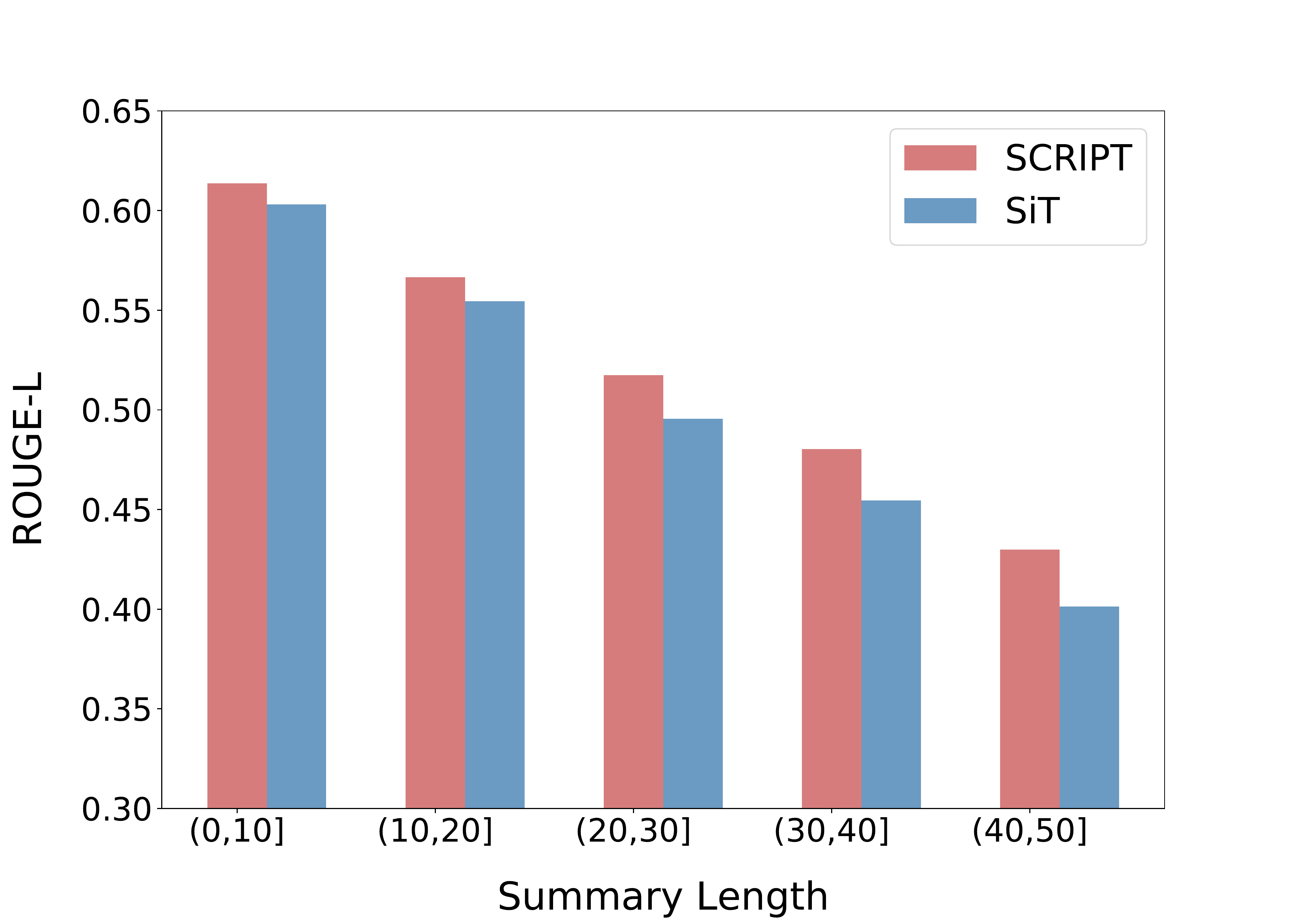}
    }
    \caption{Experimental results of our proposed method and SiT on Java dataset of different summary lengths.}
    \label{fig5}
\end{figure}

The purpose of this discussion is to study how the code length affects the representations of code, and how the summary length affects the performance of text generation.

We calculate the BLEU and ROUGE-L scores with varying the length of source code and summary. Figure~\ref{fig4} and Figure~\ref{fig5} show the performance of \tool and SiT on Java benchmark when varying code lengths and summary lengths, respectively.

As Figure~\ref{fig4} illustrates, the average BLEU and ROUGE-L scores tend to be lower for longer source code. When the code length is within the smallest two intervals, both models achieve higher BLEU scores. This means that the longer the code, the harder for the models to learn the code representations. For all code length intervals, the averages of two metrics scores of \tool outperform the averages of SiT. For the performance gap between \tool and SiT in each interval, we can find the gap of last interval is the largest. This shows AST relative positional information is helpful for model to capture long-range dependencies of code, since it closes the distance between the tokens that are far in sequential input.

As shown in Figure~\ref{fig5}, for summary of different lengths, we observe a similar trend with code of different lengths. In addition, we also find that two models are more sensitive to summary of different lengths than code of different lengths. The reason is that the metrics scores of summary length drop more significantly than code length. The observation indicates that the longer the summary, the harder to generate it completely. However, \tool still performs well when we need to generate summaries consisting of more than 30 words. And the results demonstrate that AST relative positional information also help model to generate longer summaries.

\subsection{Analysis of Structural Relative Position Learning}\label{subsec:vis}

\begin{figure}
    \centering
    \subfigure[SiT]{
    \includegraphics[width=7.47cm]{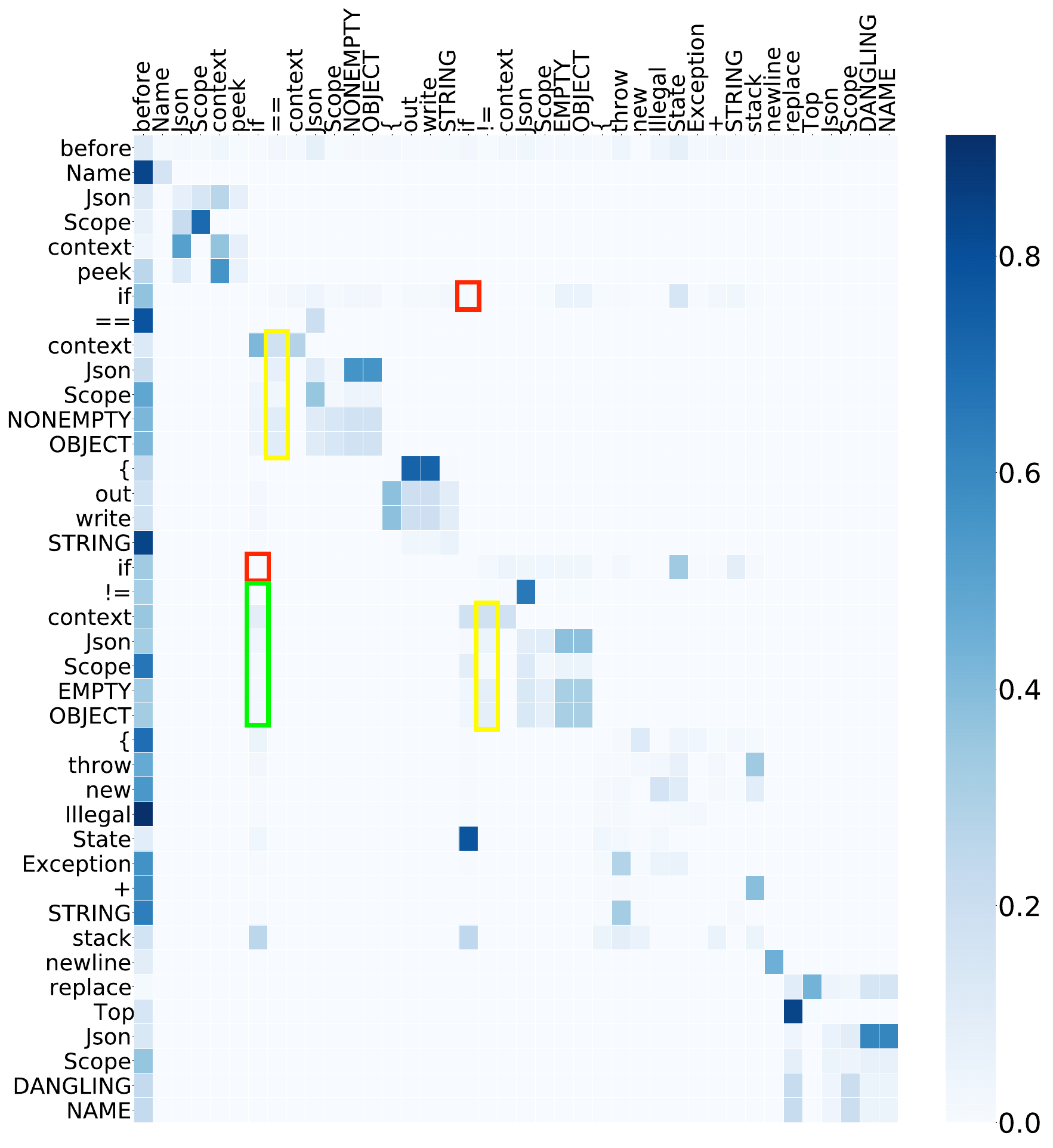}
    }
    \quad
    \subfigure[\tool]{
    \includegraphics[width=7.47cm]{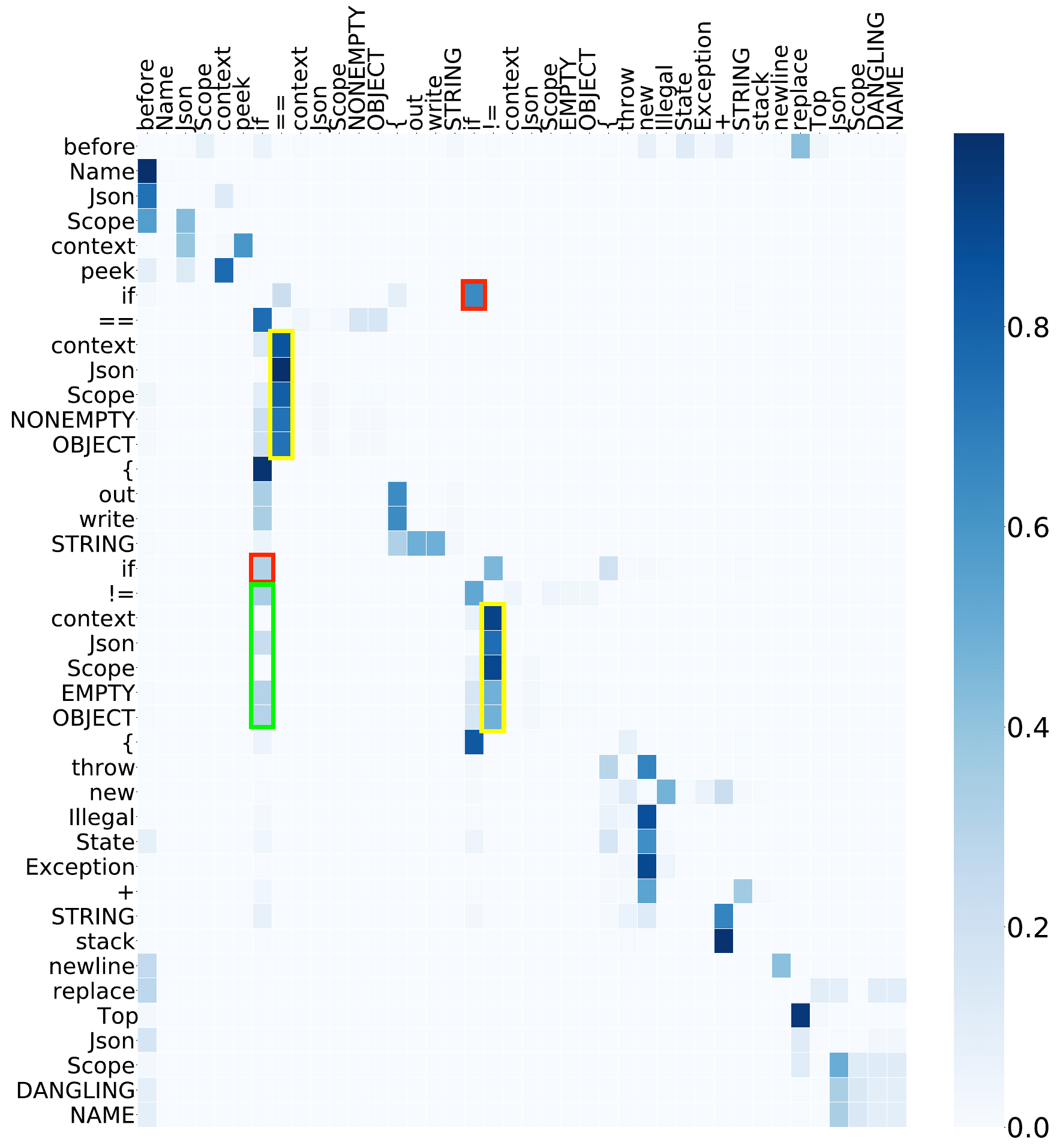}
    }
    \caption{Self-attention visualization of the fourth head in the 6-th Transformer encoder layer.}
    \label{fig6}
\end{figure}

Here, we want to figure out whether \tool is capable of learning structural relative positional knowledge. We visualize the self-attention of the first head in the sixth Transformer encoder layer, which provides an in-depth example of contribution of \tool to learn structural relative position information. The attention visualization of \tool and SiT as shown in Figure~\ref{fig6}, the corresponding source code, summaries and AST are posted in Figure~\ref{fig1}. 

First, we extract three groups of attention, and marked them with boxes of red, green and yellow respectively. Then we discuss the role of each group respectively as follows:
\begin{itemize}
    \item \textbf{Red group} represents the attention between the two underlined ``if" tokens. As shown in Figure~\ref{fig1}(a) and Figure~\ref{fig1}(b), the distance between the two underlined ``if" tokens is far in the sequential input, but they are close in the corresponding AST (only 1 hop). For SiT, the attention between the two tokens is very little. Thus, SiT does not well capture the structural relations between the two underlined ``if" tokens. For \tool, however, the attention between the two tokens is enhanced significantly. Thus, \tool can capture the structural dependencies between the two tokens.
    \item \textbf{Green group} refers to the attention between the first ``if" token and the conditional expression of the second ``if" statement. Similarly, \tool also enhances the green group of attention. The attention further help \tool to learn the whole semantics correlation between the two ``if" statements.
    \item \textbf{Yellow group} indicates the attention among the binary operator and corresponding variables in each conditional expression of ``if" statement. \tool markedly enhances the attention as well. The augmented attention help \tool to well identify the conditional differences between the two ``if" statements. It is meaningful for \tool to learn the trigger conditions of the two ``if" statements.
\end{itemize}

In short, three groups of attention jointly help \tool to well learn when each ``if" statement is executed, and what they do. Therefore, \tool can predict relatively complete code summary.

\subsection{Threats to Validity}
There are three main threats to the validity of our approach as follows: 
\begin{itemize}
    \item The primary threat to validity for this paper is the datasets we use. First, there may exist some poor-written summaries of source code, such as unrelated and unreadable summaries. Second, we do not validate the effectiveness of \tool on the datasets of other programming languages, such as C and C\#. We will try to build high-quality datasets in more different languages for evaluation in future work.
    \item The automated metrics we use pose another potential threat to validity, since they may not be representative of human judgement. BLEU, ROUGE, and METEOR metrics can calculate the textual difference between generated summaries and references. However, in some cases, the model may produce valid summaries that do not align with the ground truth, i.e., the metrics cannot truly reflect the semantic similarity. Thus, we perform manual evaluations additionally for alleviating the threat.
    \item Another threat to validity is that we do not consider other structural relative position information, such as Data Flow Graphs (DFGs) and Control Flow Graphs (CFGs). We will try to utilize the information for better learning code semantics in future work.
\end{itemize}
\section{Related Works}
In this section, we introduce the related works of source code summarization. We first collect some relevant and representative literature in the field of code-to-text over the last decade. Second, we identify their main features including IR-based, Neural Network based, utilizing the structure information and adopting pre-training in Table~\ref{tab:1}. Finally we briefly categorize the previous related works into five categories and present them one by one.

\begin{table}[htb]
    \centering
    \begin{tabular}{lcccc}
    \bottomrule
    & IR & NN & SI & PT \\
    \hline
    Haiduc \textit{et al.} (2010) \cite{haiduc2010use} & \checkmark & & & \\
    DeLucia \textit{et al.} (2012) \cite{de2012using} & \checkmark & & & \\
    Rodeghero \textit{et al.} (2014) \cite{rodeghero2014improving} & \checkmark & & & \\
    Iyer \textit{et al.} (2016) \cite{iyer2016summarizing} & & \checkmark & & \\
    Loyola \textit{et al.} (2017) \cite{loyola2017neural} & & \checkmark & & \\
    Hu \textit{et al.} \cite{hu2018summarizing} (2018) & & \checkmark & & \\
    Panthaplackel \textit{et al.} \cite{panthaplackel2020learning} (2020) & & \checkmark & & \\
    Hu \textit{et al.} \cite{hu2018deep} (2018) & & \checkmark & \checkmark & \\
    Liang \textit{et al.} \cite{liang2018automatic} (2018) & & \checkmark & \checkmark & \\
    LeClair \textit{et al.} \cite{leclair2019neural} (2019) & & \checkmark & \checkmark & \\
    Shido \textit{et al.} \cite{shido2019automatic} (2019) & & \checkmark & \checkmark & \\
    Liu \textit{et al.} \cite{liu2020automatic} (2020) & & \checkmark & \checkmark & \\
    LeClair \textit{et al.} \cite{leclair2020improved} (2020) & & \checkmark & \checkmark & \\
    Liu \textit{et al.} \cite{liu2020retrieval} (2020) & \checkmark & \checkmark & \checkmark & \\
    Zhang \textit{et al.} \cite{zhang2020retrieval} (2020) & \checkmark & \checkmark &  & \\
    Ahmad \textit{et al.} \cite{ahmad2020transformer} (2020) & & \checkmark & & \\
    Dowdell \textit{et al.} \cite{dowdell2020language} (2020) & & \checkmark & & \\
    Wu \textit{et al.} \cite{wu2020code} (2021) & & \checkmark & \checkmark & \\
    Feng \textit{et al.} \cite{feng2020codebert} (2020) & & \checkmark & & \checkmark \\
    Ahmad \textit{et al.} \cite{ahmad2021unified} (2021) & & \checkmark & & \checkmark \\
    Guo \textit{et al.} \cite{guo2020graphcodebert} (2020) & & \checkmark & \checkmark & \checkmark \\
    \bottomrule
    \end{tabular}
    \caption{Collection of related and representative publications in the last decade. Column ``IR" indicates that the approach uses Information Retrieval technology. Column ``NN" indicates that the approach is based on Neural Network. Column ``SI" indicates that the approach introduces structural information of source code. Column ``PT" indicates that the approach adopts pre-training paradigm.}
    \label{tab:1}
\end{table}

\subsection{IR-based Approaches}
In earlier studies \cite{haiduc2010use,eddy2013evaluating,wong2015clocom} for code summarization focus on using information retrieval to retrieve keywords in source code, and synthesize them into one summary. Recently, Zhang \textit{et al.} \cite{zhang2020retrieval} propose Rencos, a retrieval-based Neural Machine Translation model for code summarization, which can take advantage of both neural and retrieval techniques. And Wei \textit{et al.} \cite{wei2020retrieve} use the existing summaries of similar code snippets as exemplars to guide summary generation. While such type approaches can effectively leverage the existing terms of the original code or summaries of similar code snippets, they would generate summaries with poor readability.
\subsection{RNN-based Approaches}
With the rapid development of deep learning, researchers gradually are interested in using neural networks for the task of source code summarization. Iyer \textit{et al.} \cite{iyer2016summarizing} propose to leverage Long Short Term Memory (LSTM) networks with attention mechanism to generate summaries for source code snippets. In addition, Wei \textit{et al.} \cite{wei2019code}, \cite{ye2020leveraging} and Wan \textit{et al.} \cite{wan2018improving} prove the effectiveness to improve model by using dual learning and reinforce learning, respectively.
\subsection{Structure-based Approaches}
To learn syntax information of source code better, recent works on code summarization pay more and more attention to encode structural information, such as ASTs. AST is more widely used than DFGs or CFGs as it contains more complete structural information. Hu \textit{et al.} \cite{hu2018deep} propose a Structure-based Traversal (SBT) method to flatten the AST into a sequence as inputs and train with LSTM. LeClair \textit{et al.} \cite{leclair2019neural} treat both code text representation and SBT representation separately to learn better code semantics. Shido \textit{et al.} \cite{shido2019automatic} and Lin \textit{et al.} \cite{lin2021improving} used Tree-LSTM to model structure-style inputs like ASTs. Unlike Seq2Seq models, Liu \textit{et al.} \cite{liu2020automatic,liu2020retrieval} and LeClair \textit{et al.} \cite{leclair2020improved} innovatively utilize Graph Neural Networks (GNNs) to model structural representations of source code.
\subsection{Transformer-based Approaches}
As we all known that RNN-based models have a drawback of modeling long-range dependencies. However, Transformer overcomes this problem by using its unique multi-head self-attention mechanism. Ahmad \textit{et al.} \cite{ahmad2020transformer} propose a Transformer-based model with relative positional encoding and copy mechanism to capture long-range dependencies. Wu \textit{et al.} \cite{wu2020code} propose the structure-induced Transformer, which introduces multi-view graph matrix into self-attention mechanism process.
\subsection{Pre-training-based Approaches}
With the pre-training paradigm is widely used in NLP, increasingly more source code pre-training models are proposed. Feng \textit{et al.} \cite{feng2020codebert} propose CodeBERT which used masked language model and replaced token detection as pre-training tasks. To model structural representations of source code, Guo \textit{et al.} \cite{guo2020graphcodebert} propose GraphCodeBert which introduced two structure-aware pre-training tasks. Although the pre-training-based approaches achieve powerful performance on downstream tasks, our work is orthogonal to them.
\section{Conclusion and Future work}
In this paper, we propose a Transformer-based neural approach named \tool, which can well learn structural semantics of code for source code summarization. We build \tool upon two novel Transformers, and both of them utilize AST relative positions to augment the structural correlations between code tokens. We have evaluated the effectiveness of \tool through extensive experiments and the results show that our proposed approach outperforms the competitive baselines. In the future, we want to evaluate the effectiveness of our proposed method \tool by expanding it to more code-to-text datasets. And we plan to transfer our approach to other tasks related to source code representation learning, such as source code search.
\section*{Acknowledgment}
This paper has been supported by the National Natural Science Foundation of China (No. 62002084), Shenzhen Science and Technology Innovation Commission (No. JCYJ20200109113403826) and Stable support plan for colleges and universities in Shenzhen (No. GXWD20201230155427003-20200730101839009).

\bibliographystyle{IEEEtran}
\bibliography{reference}

\end{document}